\documentclass{Interspeech2024}
\usepackage{graphicx}
\usepackage{subcaption} 
\usepackage{relsize}
\usepackage{tipa}
\usepackage{xcolor}
\usepackage[most]{tcolorbox}
\usepackage{mathptmx}
\RequirePackage{fix-cm}

\interspeechcameraready

\title{Beyond Binary: Multiclass Paraphasia Detection with Generative Pretrained Transformers and End-to-End Models}

\name[affiliation={1}]{Matthew}{Perez}
\name[affiliation={1}]{Aneesha}{Sampath}
\name[affiliation={1}]{Minxue}{Niu}
\name[affiliation={1}]{Emily}{Mower Provost}

\address{
  $^1$University of Michigan, USA
  }
\email{mkperez@umich.edu, saneesha@umich.edu, sandymn@umich.edu, emilykmp@umich.edu}


\begin{document}

\maketitle

\begin{abstract}
    Aphasia is a language disorder that can lead to speech errors known as paraphasias, which involve the misuse, substitution, or invention of words. Automatic paraphasia detection can help those with Aphasia by facilitating clinical assessment and treatment planning options. However, most automatic paraphasia detection works have focused solely on binary detection, which involves recognizing only the presence or absence of a paraphasia. Multiclass paraphasia detection represents an unexplored area of research that focuses on identifying multiple types of paraphasias and where they occur in a given speech segment. We present novel approaches that use a generative pretrained transformer (GPT) to identify paraphasias from transcripts as well as two end-to-end approaches that focus on modeling both automatic speech recognition (ASR) and paraphasia classification as multiple sequences vs. a single sequence. We demonstrate that a single sequence model outperforms GPT baselines for multiclass paraphasia detection.
    
\end{abstract}


\section{Introduction}

Aphasia is a common language disorder that occurs as a result of damage to the brain and can ultimately impair the communication abilities (both expressive and receptive) of an individual. Aphasia affects over two million people in the United States and nearly 225,000 acquire Aphasia each year following a medical event such as a traumatic brain injury or stroke~\cite{aphasiaweb}.
Aphasia can manifest in a variety of ways, but generally, persons with Aphasia (PWAs) struggle with verbal communication and in some cases produce specific speech errors known as paraphasias.

There are several types of paraphasic errors. In this work, we focus specifically on phonemic, neologistic, and semantic paraphasias~\cite{helm2004manual,SALING200731}.
\begin{tcolorbox}
\begin{itemize}
    \item \emph{\textbf{phonemic}} paraphasias involve substituting, omitting, or rearranging phonemes (i.e., `zut' for `shut')
    \item \emph{\textbf{neologistic}} paraphasias involve substituting a nonsensical word (i.e.,  `flibber' for `bottle')
    \item \emph{\textbf{semantic}} paraphasias involve substituting a semantically related word (i.e.,  `bed' for `desk')
\end{itemize}
\end{tcolorbox}
Clinical research has highlighted the impact that accurate paraphasia detection plays in predicting recovery patterns and guiding treatment planning~\cite{fergadiotis2016algorithmic,mckinnon2018types}.
In clinical settings, automated tools for detecting paraphasias in an individual's speech can ultimately allow for more efficient and consistent assessment procedures.
Additionally, for supplementary treatment options such as remote, self-directed speech therapy (via smartphone), automatically identifying paraphasic errors is critical in providing constructive feedback to the user~\cite{ballard2019feasibility,kurland2014ipractice}.

Previous automatic paraphasia detection work has focused on identifying paraphasias from single-word elicitation tasks with manual transcriptions~\cite{fergadiotis2016algorithmic,casilio2023paralg,salem2023refining}. These works have limited applications, mainly in clinical settings. For applications with continuous or unsegmented speech, paraphasia detection includes identifying where in the given sequence a paraphasia occurs. Some previous works that have focused on continuous speech have treated paraphasia detection as a binary task~\cite{le2017paraphasia,pai2020unsupervised,perez2023seq2seq}. However, these works are restricted to learning the presence or absence of paraphasic errors rather than learning to differentiate between paraphasia types. For remote speech therapy applications that process continuous speech, models that focus on multiclass classification are needed to characterize these different types of paraphasias and where they occur in a given utterance.

We present the first work into automatic multiclass paraphasia detection for continuous speech. We investigate several methods for automatic paraphasia classification, which include using a generative pretrained transformer to classify paraphasias from ASR transcripts (ASR+GPT), a single-sequence (single-seq) model where both ASR and paraphasia classification tasks are learned within the same sequence, and a multi-sequence (multi-seq) model where paraphasia classification and ASR are learned as separate sequences but jointly optimized with multitask learning. The research goals of this work are:
\begin{itemize}
    \item To investigate the utility of an off-the-shelf GPT model for paraphasia classification using imperfect (ASR) or perfect (oracle) transcripts.
    \item Explore single-seq and multi-seq models for word-level paraphasia classification.
    \item Analyze performance across different paraphasia classes.
\end{itemize}
Our findings demonstrate that GPT can be used to detect paraphasias with aphasic speech transcripts. However, we note that the single-seq model outperforms GPT for multiclass paraphasia detection, specifically for phonemic and neologistic paraphasias. Lastly, we discuss some limitations of the presented approaches for semantic paraphasia classification.

\begin{table*}
\vspace{-4mm}
\caption{Text Pre-Processing: CHAT transcriptions are processed to Oracle transcripts. Examples for each model output is also shown. \textcolor{blue}{Blue} indicates paraphasic words, \textcolor{red}{red} indicates paraphasic labels}
\vspace{-2mm}
    \centering
    \small
    \setlength{\tabcolsep}{3pt}
    \begin{tabular}{|l|l|}
        \toprule
         CHAT Transcripts & \text{aphasia \textcolor{blue}{f\textepsilon kts@u} [: affects] \textcolor{red}{[* p]} my language not my 
         \textcolor{blue}{d\textipa{I}t\textipa{I}k\textschwa lt@u} [: intelligence] \textcolor{red}{[* n]}} \\
         \midrule
         Oracle & aphasia \textcolor{blue}{fekts} \textcolor{red}{[p]} my language not my \textcolor{blue}{ditikalt} \textcolor{red}{[n]}  \\
         \midrule
         \multicolumn{2}{|c|}{Model Output} \\
         \midrule
         ASR+GPT & aphasia [c] fekts [p] my [c] language [c] not [c] my [c] ditikalt [n] \\
         \midrule
         single-seq & aphasia fekts [p] my language not my ditikalt [n]  \\
         \midrule
         multi-seq & ASR: aphasia fekts my language not my ditikalt \\
          & Para: [c]     [p]  [c]   [c]  [c]  [c]  [n] \\
         \bottomrule
    \end{tabular}
    \vspace{-3mm}
    \label{tab:processing}
\end{table*}

\section{Related Work}
One of the first works that explored statistical models for paraphasia detection was by Fergadotis et al., which used separate classification models to perform binary paraphasia detection in a one-vs-rest fashion~\cite{fergadiotis2016algorithmic}.
This work was focused on the Moss Aphasia Psycholinguistics Project Database (MAPPD), which contains transcribed (text-input) single-word responses with paraphasia labels for each word~\cite{mirman2010large}. The classifiers use linguistic features like word2vec or semantic similarity for paraphasia classification.
One limitation of this work is that paraphasia classification is performed by separate classifiers, which lacks the specificity to differentiate between different paraphasias.
More recent work~\cite{casilio2023paralg,salem2023refining} has addressed these concerns by presenting a unified model for multiclass paraphasia detection on MAPPD where a multiclass decision tree based on the binary classifiers presented in~\cite{fergadiotis2016algorithmic} is used to perform multiclass paraphasia detection.
A broader limitation of the works above is that MAPPD, is constrained to single-word responses that are transcribed (i.e., contain no audio data). This is useful in clinical applications where manual transcription for targeted tasks can be attained. However, for unconstrained speech applications such as remote speech therapy, automatic paraphasia detection models must be able to handle unsegmented or continuous speech data.
In this work, we investigate multiclass paraphasia detection for continuous, read speech from the AphasiaBank corpus~\cite{macwhinney2014childes}. 
The paraphasia classification in continuous speech is complicated by temporal challenges, which involve not just identifying the different types of paraphasias but also the specific points in the utterance where they occur.

Some additional works have focused on paraphasia detection in continuous speech.
Work by Le et al. showed that a fully automatic pipeline composed of ASR and a logistic regression model for binary paraphasia detection can detect phonemic and neologistic paraphasias in continuous speech~\cite{le2017paraphasia}. Work by Pai et al. demonstrated that density-based clustering with manual segmentation can yield improved features for binary paraphasia detection~\cite{pai2020unsupervised}. Lastly, work by Perez et al. focused on automatic paraphasia detection using end-to-end models and showed that multitask learning improved performance over existing automated approaches~\cite{perez2023seq2seq}.
These works are limited in that they focus on binary paraphasia detection in continuous speech and only detect phonemic and neologistic paraphasias. For remote applications, multiclass paraphasia detection models that can process continuous speech are needed to locate when paraphasias occur and accurately differentiate between paraphasia types.
In this work, we present several methods for performing multiclass paraphasia detection in continuous speech that include phonemic, neologistic, and semantic paraphasias.

Outside of paraphasia detection, Omachi et al. has shown that a sequence-to-sequence model can learn to produce ASR and linguistic annotations in a single sequence~\cite{omachi2021end}. In our work, we explore a similar framework that learns to produce ASR and paraphasia annotations from aphasic speech.

\section{Data}
We use two English datasets from the AphasiaBank corpus. The first is the Protocol dataset, which contains both aphasic and healthy speakers performing a series of tasks following the Western Aphasia Battery~\cite{kertesz2022western}. The Protocol dataset is first used to train our models as this dataset is the larger of the two, containing roughly 100 hours of speech. The second dataset is the Fridriksson subset, which contains roughly three hours of speech, where PWAs were asked to read a series of scripts. The Fridriksson dataset contains a higher distribution of paraphasias compared to the Protocol dataset, with phonemic, neologistic, and semantic paraphasias representing roughly 13\%, 7\%, and 3\% of the dataset, respectively. Similar to previous works, we use the Fridriksson dataset for finetuning and evaluation.

All utterances were manually transcribed in the CHAT format and included timestamps for both participant and interviewer speech segments~\cite{macwhinney2014childes}. 
We isolated participant speech and discarded utterances that were labeled unintelligible or containing overlapping speech between the participant and clinician.
The CHAT transcription is processed by removing punctuation and reducing all characters to lowercase. For non-word phonological errors transcribed in the International Phonetic Alphabet (IPA) format, we convert these to pseudo-words following previous works~\cite{le2017paraphasia} by mapping each IPA pronunciation to a sequence of phones and then converting phones to graphemes.
Lastly, we label all phonemic, neologistic, and semantic paraphasias, indicated by [* p], [* n], and [* s] in the CHAT transcriptions with [p], [n], and [s] respectively.
The final processed transcript can be seen in the Oracle section of Table~\ref{tab:processing}.

\begin{figure*}[t]
    \centering
    \vspace{-3mm}
    \begin{subfigure}[b]{0.49\linewidth} 
        \includegraphics[width=\linewidth]{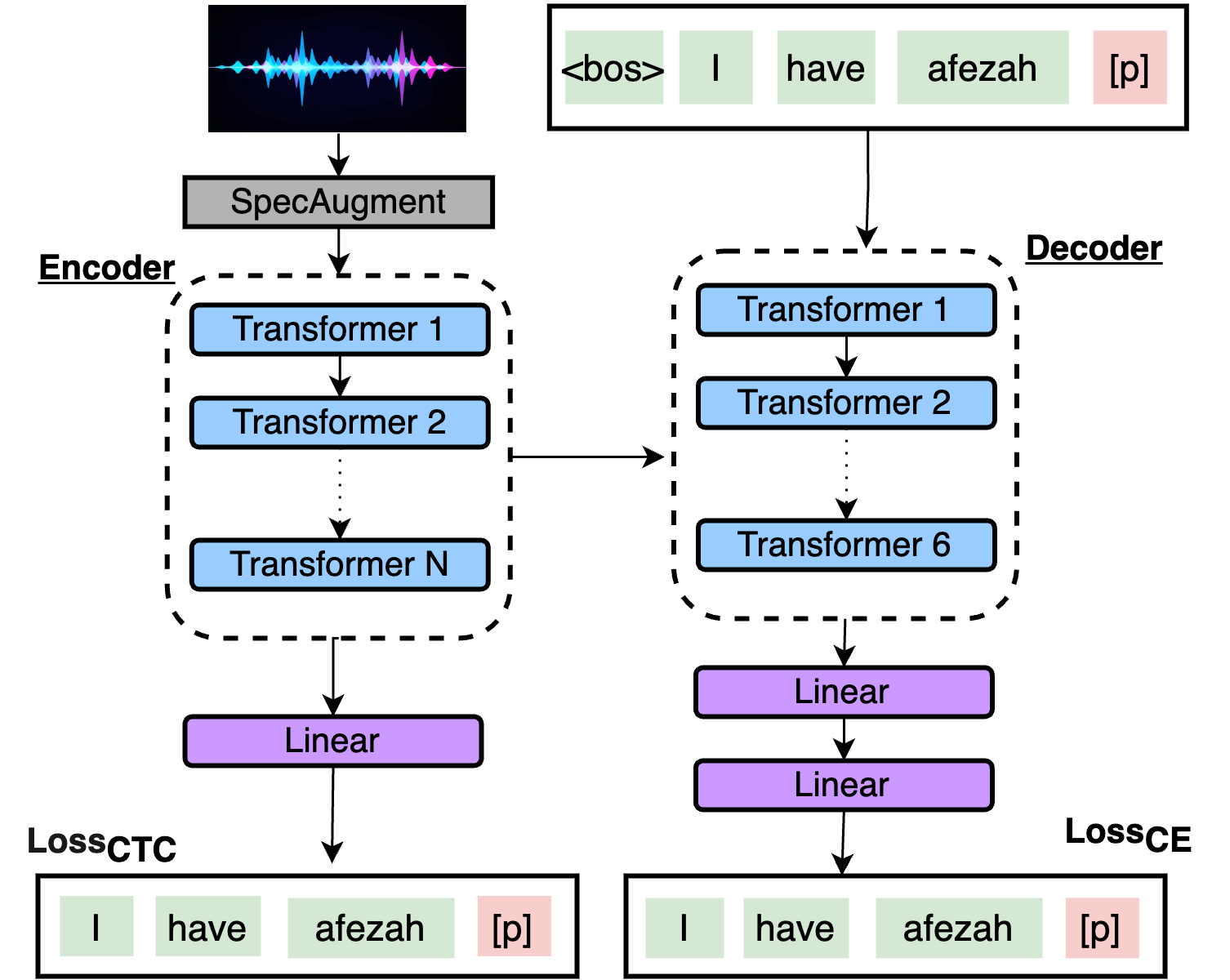}
        \caption{ASR / single-seq}
        \label{fig:1a}
    \end{subfigure}
    \hfill
    \begin{subfigure}[b]{0.49\linewidth} 
        \includegraphics[width=\linewidth]{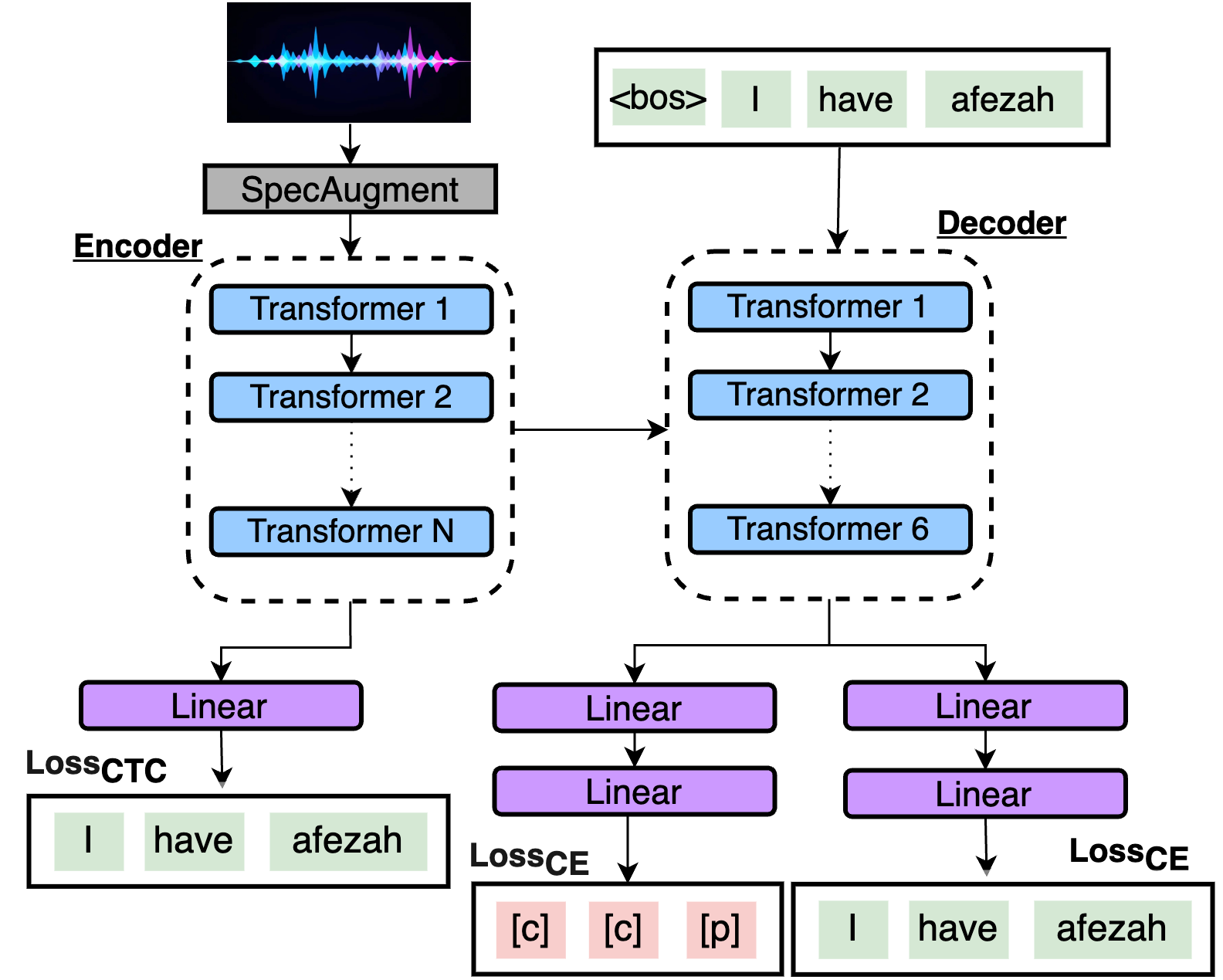}
        \caption{multi-seq}
        \label{fig:1b}
    \end{subfigure}
    \vspace{-2mm}
    \caption{Paraphasia Classification Models.}
    \vspace{-3mm}
    \label{fig:models}
\end{figure*}

\section{Methods}

\subsection{Transcript+GPT}
We explore a pipeline that consists of using speech transcripts and a generative pretrained transformer (GPT) model to classify paraphasias. We perform in-context learning on openAI's GPT-3.5-turbo and GPT-4 by conditioning these models with task instructions and an example. In-context learning is an effective approach for tuning GPT output for a wide variety of NLP tasks without having to update model parameters~\cite{perez2021true,lu2021fantastically}.
We experiment with two approaches for transcript generation, the first uses an in-domain ASR model and the second uses the manual (oracle) transcripts.
\subsubsection{ASR+GPT}
\label{sec:ASR+GPT}
The ASR model shown in Figure~\ref{fig:1a} is inspired by previous work~\cite{perez2023seq2seq}. We use both conntectionist temporal classification (CTC) loss~\cite{graves2006connectionist} and cross-entropy (CE) loss. The model is optimized using a joint CTC-attention loss criterion~\cite{kim2017joint}.
We also use SpecAugment~\cite{park2019specaugment} to resample utterances at different time perturbations
using rates of [0.8, 0.9, 0.95, 1.0, 1.05, 1.1, 1.2] which has been shown to be effective for disordered speech recognition~\cite{green2021automatic,perez2023seq2seq}.

\subsubsection{Oracle+GPT}
We explore the effect that ASR errors have on the performance of GPT's paraphasia classification. We assume perfect ASR transcription (i.e., an oracle) by using the ground truth transcriptions as input to the GPT model. Comparing ASR+GPT and Oracle+GPT will highlight the performance gap in paraphasia classification introduced by inaccurate transcription.

\subsection{Single-Seq}
The single sequence model (Figure~\ref{fig:1a}) has the same architecture as the ASR model but the tokenizer has three added special tokens representing [p], [n], and [s] paraphasia labels. 
The seq2seq model learns to predict subword tokens (graphemes/morphemes) related to ASR and the special tokens related to paraphasia classification.
The single-seq model learns to predict paraphasia labels after transcribing a paraphasic word.
By treating paraphasias as a separate token, the model can learn contextual dependencies across subword and paraphasia tokens.
The loss function for this model can be seen in equation~\ref{eq:SS_loss}, where $z_t$ represents either a subword token or a paraphasia token at time $t$, $M$ is the combined length of subword tokens and paraphasia labels in the utterance, and $x$ is the audio representation from the encoder.
\begin{equation}
\vspace{-2mm}
\begin{aligned}
    \mathcal{L} = -\sum_{t=0}^{M}\log P(z_t|z_0,...,z_{t-1},x)
\end{aligned}
\label{eq:SS_loss}
\vspace{-1mm}
\end{equation}

\subsection{Multi-Seq}
The multi-sequence model (shown in Figure~\ref{fig:1b}) has separate ASR and paraphasia classification heads to produce independent sequences. The paraphasia classification head consists of two fully connected layers with the final layer having an output size of 4 corresponding to [c] (non-paraphasia), [p], [n], and [s].  In this setup, paraphasia classification is performed at the subword level.
We use a weighted paraphasia classification loss based on the inverse of the paraphasia class count. The total loss combines both the paraphasia classification loss and the ASR loss seen in equations~\ref{eq:multi-seq_loss} and \ref{eq:multi-seq_loss_tot}, where $y_t$ and $p_t$ represent the ASR token and paraphasia class predictions at time $t$ and $T$ is the total length of the subword tokens in the utterance. We conduct a hyperparameter sweep for $\alpha$ ranging from 0.3 to 0.7 and select the optimal value based on development set performance.

\vspace{-2mm}
\begin{equation}
\begin{aligned}
    \mathcal{L}_{asr} = -\sum_{t=0}^{T}\log P(y_t|y_0,...,y_{t-1},x) \\
    \mathcal{L}_{para} = -\sum_{t=0}^{T}\log P(p_t|y_0,...,y_{t-1},x) \\
\end{aligned}
\label{eq:multi-seq_loss}
\end{equation}
\begin{equation}
\begin{aligned}
\mathcal{L} = \alpha\mathcal{L}_{asr} + (1-\alpha)\mathcal{L}_{para}
\end{aligned}
\label{eq:multi-seq_loss_tot}
\end{equation}

The ASR and paraphasia classification heads use the same decoder representation, which ensures alignment at the subword level.
For training, we map paraphasia labels to the subword level by assigning the paraphasia label of a given word to all its constituent subword tokens. 
At inference, we attain word-level paraphasia predictions by using the majority paraphasia class across all subwords in a given word. During decoding, beam pruning is based on ASR output scores only.

\section{Experiment Setup}
All our models were built using the SpeechBrain toolkit~\cite{ravanelli2021speechbrain} and consist of an encoder and a decoder with 24- and 6-transformer layers respectively. We initialize the encoder parameters with a pretrained HuBERT model\footnote{https://huggingface.co/facebook/hubert-large-ls960-ft}. We use a tokenizer with a vocabulary size of 500 that performs reductions using byte-pair encoding (BPE) and use learning rate annealing with a factor of 0.8 based on previous work~\cite{perez2023seq2seq}. Due to hardware constraints we use a batch size of 4 with a gradient accumulation of 4. Our code, model output samples, and additional supplementary information related to data splits, hyperparameters, and prompts for in-context learning can be found on github\footnote{https://github.com/chailab-umich/BeyondBinary-ParaphasiaDetection}.

\subsection{Standardizing Model Output}
Prior to evaluation, we first standardize the output from each model to take the format of ASR+GPT in Table~\ref{tab:processing}. Specifically, we pre-process all model outputs to the form $\hat{Y}=\hat{y_0},\hat{p_0},...,\hat{y_n},\hat{p_n}$, where $\hat{y_n}$ and $\hat{p_n}$ represent the predicted word and paraphasia label at $n$-index. 

\subsection{WER Metrics}

\textbf{Word-error-rate (WER)} measures ASR performance, where $\hat{Y}=\hat{y_0},...,\hat{y_n}$ and $Y=y_0,...,y_m$ represent the predicted and ground truth sequences used in the evaluation.
\textbf{Augmented-word-error-rate (AWER)} is used to measure both ASR and paraphasia classification performance, where WER is computed using $\hat{Y}=\hat{y_0},\hat{p_0},...,\hat{y_n},\hat{p_n}$ and $Y=y_0,p_0,...,y_m,p_m$, which represent the predicted and ground truth sequences. This metric is based on previous work~\cite{le2017paraphasia,perez2023seq2seq}.



We use the aggregated labels and predictions across all test folds and compute a single metric similar to~\cite{le2017paraphasia,pai2020unsupervised,perez2023seq2seq}.
We perform statistical significance testing for WER and AWER metrics using a bootstrap estimate, which is commonly used for determining statistical significance across ASR systems~\cite{bisani2004bootstrap}. For each comparison, we perform 1000 iterations using a batch size of 100 and adopt a 95\% confidence threshold for declaring statistical significance.

\begin{table}[t]
\caption{Word-level results aggregated over all folds with best results in bold (oracle transcripts not included). $\dagger$ indicates statistical significance over the baseline GPT-4 approach. For all metrics, lower values indicate better performance.} 
\vspace{-1mm}
    \centering
    \smaller
    \setlength{\tabcolsep}{4pt}
    \begin{tabular}{l|c|c|c|c|c|c|c}
    \toprule
    \multicolumn{4}{c}{} & \multicolumn{4}{|c}{TD-multiclass}\\
    \midrule
    Models & WER & AWER & TD-bin & [p] & [n] & [s] & all \\ 
    \midrule
    \multicolumn{8}{l}{\emph{ASR Transcripts (Baseline)}} \\
    \midrule
    GPT-3.5             & 38.6 & 32.9 & 1.17 & 1.02 & 0.67 & 0.30 & 1.99 \\
    GPT-4               & 38.6 & \textbf{32.7} & 0.68 & 0.86 & 0.73 & \textbf{0.29} & 1.88 \\
    \midrule
    \multicolumn{8}{l}{\emph{Proposed}} \\
    \midrule
    Single-Seq          & \textbf{37.6} & 32.8 & \textbf{0.63}$\dagger$ & \textbf{0.76}$\dagger$ & \textbf{0.45}$\dagger$ & 0.31 & \textbf{1.51}$\dagger$ \\
    Multi-Seq               & 44.8 & 42.9 & 0.86& 0.90 & 0.72  & 0.53 & 2.15 \\
    \midrule
    \multicolumn{8}{l}{\emph{Oracle Transcripts}} \\
    \midrule
    GPT-3.5          & --  & 10.6 & 0.82 & 0.97 & 0.54 & 0.30 & 1.81 \\
    GPT-4            & --  & 7.9  & 0.25 & 0.67 & 0.44 & 0.27 & 1.38 \\
    \bottomrule
    \end{tabular}
    \vspace{-4mm}
    \label{tab:word-level}
\end{table}

\subsection{Distance Metrics}
WER metrics have the limitation of measuring accuracy at a given index, which can be misleading for evaluating paraphasia classification if there is misalignment due to ASR.
We use temporal distance (TD) to evaluate word-level paraphasia classification, which measures the proximity to the closest paraphasia class~\cite{kovacs2019evaluation,perez2023seq2seq}. 
We align the words of $Y$ and $\hat{Y}$ using the minimum edit operations, compute the TD metric using the paraphasia class labels, and normalize the resulting TD by the utterance length. 
We present several evaluations with TD.
\textbf{TD-binary} measures the proximity of detecting any paraphasic instance. \textbf{TD-multiclass} measures the proximity between specific paraphasias, we compute individual metrics for each paraphasia class (\textbf{TD-[p,n,s]}) as well as an overall TD across all classes (\textbf{TD-all}).
We present the average TD across all test utterances and perform statistical significance testing using a repeated measures ANOVA followed by a post-hoc tukey test, where statistical significance is determined by a p-value $<$ 0.05.

\subsection{Utterance-level Metrics}
We assess paraphasia classification at the utterance-level using a binary F1 score for each paraphasia type. Similar to previous work, we compute a single F1 score over all test utterances~\cite{le2017paraphasia,perez2023seq2seq}.
For a given paraphasia class, an utterance is assigned a positive class label if there is any instance of that paraphasia present.


\section{Results}
\begin{figure}
    \centering
    \includegraphics[width=\linewidth]{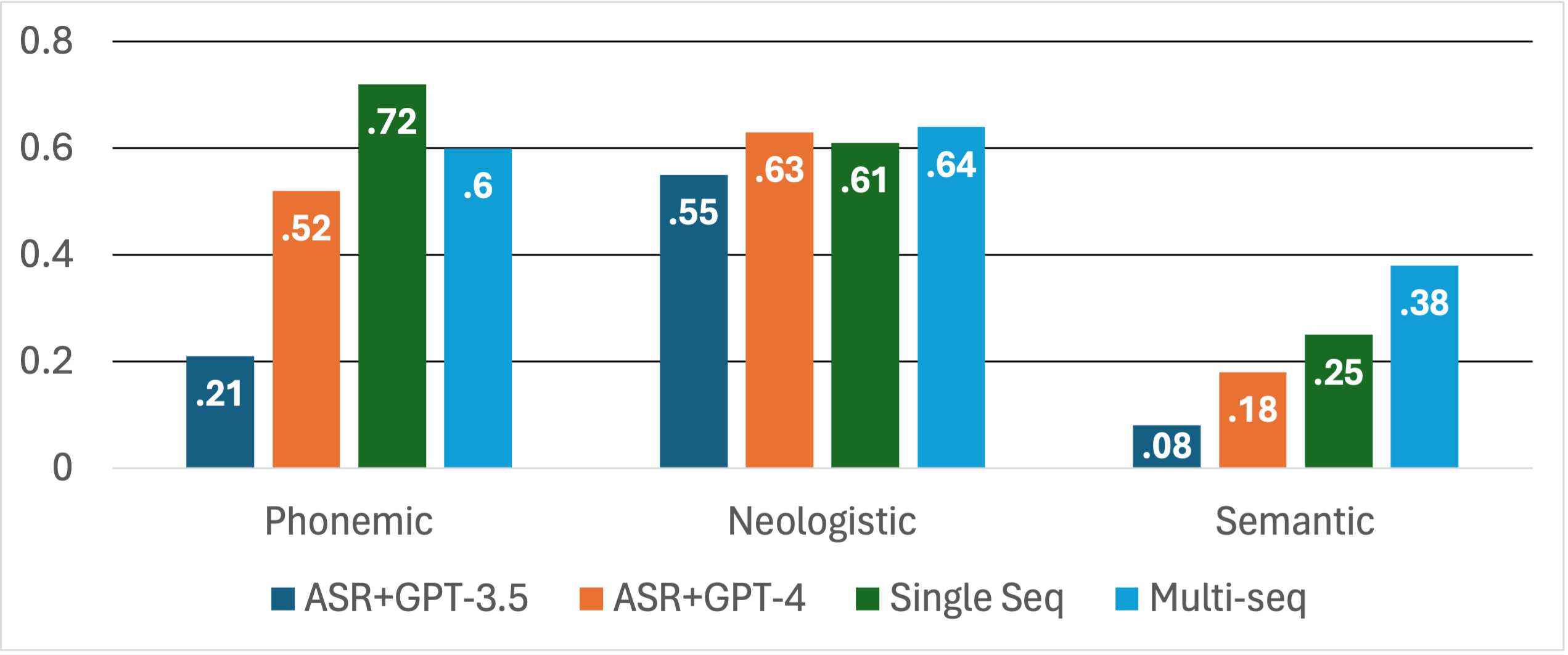}
    \vspace{-6mm}
    \caption{Utterance-level Binary F1-scores}
    \vspace{-5mm}
    \label{fig:utt-level}
\end{figure}

In Table~\ref{tab:word-level} we see that single-seq has the lowest WER while GPT-4 achieves the lowest AWER. This indicates that although single-seq may produce more accurate transcriptions, GPT-4 is more accurate when factoring in paraphasia classification.
We see that the single-seq model significantly outperforms the baseline GPT-4 for paraphasia detection (TD-binary) and multiclass classification (TD-multiclass) over most categories. 
Specifically, for phonemic paraphasias, we see that single-seq has the lowest TD-[p] and the highest F1 score (shown in Figure~\ref{fig:utt-level}) indicating its effectiveness over other models for this paraphasia type. 
For neologistic paraphasias, we observe that single-seq has the lowest TD-[n], indicating that it is the most effective at locating where neologistic paraphasias occur within an utterance.  However, the F1 score for single-seq is slightly below that of GPT-4 and multi-seq, suggesting that it may be slightly less effective (0.61 vs. 0.63 for single-sequence vs. GPT-4, respectively) at identifying the presence of a neologistic paraphasia in an utterance.

Across all approaches, classifying semantic paraphasias is a challenging task. Although TD-[s] is low compared to TD-[p] and TD-[n] this is due to the low representation of semantic paraphasias in the dataset, which ultimately results in low TD when averaged across all utterances.
This is further evident when looking at the binary F1-scores for semantic paraphasias, where all approaches struggle at predicting semantic paraphasias at the utterance-level.

The use of oracle transcripts significantly improves GPT-4's performance in detecting paraphasic instances, which is reflected by a 63\% relative performance improvement for TD-binary. For classifying all paraphasias we note a 27\% relative improvement for TD-multiclass, which highlights an existing gap for GPT-4 to differentiate between specific paraphasia classes despite having access to perfect transcription. 

\section{Conclusion}
In this work, we present several methods for classifying paraphasias, including the use of transcripts and GPT, a seq2seq model with a single output (single-seq), and a seq2seq model with multi-sequence output (multi-seq). We show the efficacy of GPT with perfect and imperfect transcripts and its limitations for certain paraphasia types. However, within the context of differentiating between all paraphasia classes, we find that the single-seq approach provides statistically significant improvement over the baseline GPT approaches that use ASR for transcription.
The performance of multi-seq warrants further investigation into methods where paraphasia predictions are made at the subword level.
Future work for improving semantic paraphasia detection should explore including conditioning models with the target script in order to provide contextual information that is relevant for identifying associative word substitutions. 


\newpage

\section{Acknowledgements}
This paper was accepted to Interspeech 2024.

\bibliographystyle{paper_kit/IEEEtran}
\bibliography{mybib}

\end{document}